\documentclass{article}
\PassOptionsToPackage{numbers, compress}{natbib}
\usepackage[final]{nips_2016}
\usepackage[utf8]{inputenc} 
\usepackage[T1]{fontenc}    
\usepackage{hyperref}       
\usepackage{url}            
\usepackage{booktabs}       
\usepackage{amsfonts}       
\usepackage{nicefrac}       
\usepackage{microtype}      
\usepackage{multirow}
\usepackage{amsmath,amssymb,bm}
\usepackage{footnote}
\usepackage{graphicx}

\makesavenoteenv{tabular}
\makesavenoteenv{table}

\title{Neural Speech Recognizer: Acoustic-to-Word LSTM Model for Large Vocabulary Speech Recognition}

\author{
  Hagen Soltau, Hank Liao, Hasim Sak\\
  Google, Inc.\\
  \texttt{\{soltau,hankliao,hasim\}@google.com} \\
}

\begin{document}

\newcommand{\argmax}{\textrm{argmax}}

\maketitle

\begin{abstract}

We present results that show it is possible to build a competitive,
greatly simplified, large vocabulary continuous speech recognition
system with whole words as acoustic units. We model the output
vocabulary of about 100,000 words directly using deep bi-directional LSTM RNNs with CTC loss.
The model is trained on 125,000 hours of semi-supervised
acoustic training data, which enables us to alleviate the data
sparsity problem for word models. We show that the CTC word
models work very well as an end-to-end all-neural speech recognition model without the use of
traditional context-dependent sub-word phone units that require a
pronunciation lexicon, and without any language model removing the need to
decode. We demonstrate that the CTC word models perform better than
a strong, more complex, state-of-the-art baseline with sub-word units.

\end{abstract}

\section{Introduction}
\label{sec:intro}

End-to-end speech recognition with neural networks has been a goal for
the machine learning and speech processing
communities~\citep{graves14icml, miao15asru, Bahdanu16icassp,
  zhang16inter, deepspeech2, Liang16icassp, chan16ica}. In the past, the best speech recognition systems have
used many complex modeling techniques for improving accuracy: for
example the use of hand-crafted feature representations, speaker or
environment adaptation with feature or affine transformations, and
context-dependent (CD) phonetic models with decision tree
clustering~\cite{Bahl91, young94} to name a few. For automatic speech
recognition, the goal is to minimize the word error rate. Therefore
it is a natural choice to use words as units for acoustic modeling and
estimate word probabilities. While some attempts had been made to
model words directly, in particular for isolated word recognition with
very limited vocabularies~\citep{lang90nn}, the dominant approach is
to model clustered CD sub-word units instead.

On the other hand, these clustered units were a necessity when data
were limited, but may now be a sub-optimal choice. Recently, the
amount of user-uploaded captions for public YouTube videos has grown
dramatically. Using powerful neural network models with large amounts
of training data can allow us to directly model words and greatly
simplify an automatic speech recognition system. It was previously
found that the combination of a LSTM RNN model's~\citep{Hochreiter:97}
memorization capacity and the ability of CTC loss~\citep{Graves:06} to
learn an alignment between acoustic input and label sequences allows a
neural network that can recognize whole words to be
trained~\citep{Sak:15b}. Although training data sparsity was an
issue, bi-directional LSTM RNN models, with a large output vocabulary
could be trained, i.e. up to 90,000 words, and obtained respectable
accuracy without doing any speech decoding, however this was still far
from the sub-word phone-based recognizer. In this paper, we show that
these techniques coupled with a larger amount of acoustic training
data enable us to build a neural speech recognizer (NSR) that can be trained
end-to-end to recognize words directly without needing to decode.

There have been many different approaches to end-to-end neural network
models for speech recognition. \citet{Liang16icassp} use an
encoder-decoder model of the conditional probability of the full word
output sequence given the input sequence. However due to the limited
capacity, they extend it with a dynamic attention scheme that makes
the attention vector a function of the internal state of the
encoder-decoder which is then added to the encoder RNN. They compare
this with explicitly increasing the capacity of the RNN. In both
cases, there is still a gap in performance with the conventional
hybrid neural network phone-based HMM recognizer. Instead of word
outputs, letters can be generated directly in a grapheme-based neural
network recognizer~\citep{Bahdanu16icassp, chan16ica}; while good
results are obtained for a large vocabulary task, they are not quite
comparable to the phone-based baseline system.



\section{Neural Speech Recognizer}
\label{sec:neural_asr}
Here, we describe the techniques that we used for building the NSR:
a single neural network model capable of accurate speech recognition with no search or decoding involved.
The NSR model has a deep LSTM RNN architecture built by stacking multiple LSTM layers.
Since the bidirectional RNN models~\citep{Schuster:97} have better accuracy and our application is offline speech recognition,
we use two LSTM layers at each depth---one operating in the forward and another operating in the
backward direction in time over the input sequence.
Both these layers are connected to both previous forward and backward layers.

We train the NSR model with the CTC loss criterion~\citep{Graves:06} which
is a sequence alignment/labeling technique with a
softmax output layer that has an additional unit for the \textit{blank} label used
to represent outputting no label at a given time.
The output label probabilities from the network define a probability
distribution over all possible labelings of input sequences including the blank
labels. The network is trained to optimize the total probability of
correct labelings for training data as estimated using the network outputs and
forward-backward algorithm.
The correct labelings for an input sequence are defined as the set of all
possible labelings of the input with the target labels in the correct sequence order
possibly with repetitions and with blank labels permitted between labels.
The CTC loss can be efficiently and easily computed using finite state transducers (FSTs) as described in~\citet{Sak:15a}
\begin{equation}
  \mathcal{L}_{CTC} = - \sum_{(\bm{x, l})} \text{ln} p(\bm{z^l|x}) = - \sum_{(\bm{x, l})} \mathcal{L}(\bm{x,z^l})
\end{equation}
where $\bm{x}$ is the input sequence of acoustic frames, $\bm{l}$ is the input label sequence (e.g. a sequence of words for the NSR model),
$\bm{z^l}$ is the lattice encoding all possible alignments of $\bm{x}$ with $\bm{l}$
which allows label repetitions possibly interleaved with \textit{blank} labels.
The probability for correct labelings $p(\bm{z^l|x})$ can be computed using the forward-backward algorithm.
The gradient of the loss function w.r.t. input activations $a_l^t$ of the softmax output layer for a training example can be computed as follows:
\begin{equation}
 \frac{\partial\mathcal{L}(\bm{x,z^l})}{\partial a_l^t} = y_l^t - \frac{1}{p(\bm{z^l|x})} \sum_{u \in \left\{u : \bm{z^l}_u = l\right\}} \alpha_{x,z^l}(t, u) \beta_{x,z^l}(t, u)
\end{equation}
where $y_l^t$ is the softmax activation for a label $l$ at time step $t$, and
$u$ represents the lattice states aligned with label $l$ at time $t$, $\alpha_{x,z^l}(t, u)$ is the forward variable representing the summed probability
of all paths in the lattice $\bm{z^l}$ starting in the initial state at time $0$ and ending in state $u$ at time $t$,
$\beta(t, u)$ is the backward variable starting
in state $u$ of the lattice at time $t$ and going to a final state.

The NSR model has a final softmax predicting word posteriors with the
number of outputs equaling the vocabulary size. Modeling words
directly can be problematic due to data sparsity, but we use a large
amount of acoustic training data to alleviate it. We experiment with both written and spoken vocabulary.
The vocabulary obtained from the training data transcripts is mapped to the spoken
forms to reduce the data sparsity further and limit label ambiguity for the spoken vocabulary experiments.
For written-to-spoken domain mapping a FST
verbalization model is used~\citep{Sak:13}. For example, ''104`` is
converted to ''one hundred four`` and ``one oh four``. Given all
possible verbalizations for an entity, the one that aligns best with
acoustic training data is chosen.

The NSR model is essentially an all-neural network speech recognizer
that does not require any beam search type of decoding. The network
takes as input mel-spaced log filterbank features.
The word posterior probabilities output
from the model can be simply used to get the recognized word sequence.
Since this word sequence is in spoken domain for the spoken vocabulary model, to get the written
forms, we also create a simple lattice by enumerating the alternate
words and blank label at each time step, and rescore this lattice with
a written-domain word language model (LM) by FST composition after composing it with the verbalizer FST.
For the written vocabulary model, we directly compose the lattice with the language model to assess the importance of language model rescoring for accuracy.

We train the models in a distributed manner using asynchronous stochastic gradient
descent (ASGD) with a large number of machines~\citep{Dean:12,Sak:14a}.
We found the word acoustic models performed better when initialized using the parameters from hidden states of phone models---the
output layer weights are randomly initialized and the weights in the initial networks are randomly initialized with a uniform (-0.04, 0.04) distribution.
For training stability, we clip the activations of memory cells to [-50, 50], and the gradients to [-1, 1] range.
We implemented an optimized native TensorFlow CPU kernel (multi\_lstm\_op) for multi-layer LSTM RNN forward pass and gradient calculations.
The multi\_lstm\_op allows us to parallelize computations across LSTM layers using pipelining and the resulting speed-up decreases the parameter staleness
in asynchronous updates and improves accuracy.

\section{Experimental Setup}
\label{sec:baseline}

YouTube is a video sharing website with over a billion users. To
improve accessibility, Google has functionality to caption YouTube
videos using automatic speech recognition technology~\citep{autocaps}.
While generated caption quality can vary, and are generally no better
than human created ones, they can be produced at scale. On the whole,
users have found them helpful: Google received a technology
breakthrough award from the US National Association of the Deaf in
2015 for automatic captioning on YouTube. For this work, we evaluate
our models on videos sampled from Google Preferred channels on
YouTube~\citep{ytpref}. The test set is comprised of 296 videos from
13 categories, with each video averaging 5 minutes in length. The
total test set duration is roughly 25 hours and 250,000
words~\citep{vitaly16inter}. As the bulk of our training data is not
supervised, an important question is how valuable this type of the
data is for training acoustic models. In all our experiments, we keep
our language model constant and use a 5-gram model with 30M N-grams
over a vocabulary of 500,000 words.

Training large, accurate neural network models for speech recognition
requires abundant data. While others have used read speech
corpora~\citep{deepspeech2, vassil15ica} or unsupervised
methods~\citep{olga14inter} to gather thousands or even tens of
thousands of hours of labeled training data, we apply an approach
first described in \citep{liao13asru} but now scaled up to build a
training set of over 125,000 hours. This ``islands of confidence''
filtering, allows us to use user-uploaded captions for labels, by
selecting only audio segments in a video where the user uploaded
caption matches the transcript produced by an ASR system constrained
to be more likely to produce N-grams found in the uploaded caption. Of
the approximately 500,000 hours of video available with English
captions, a quarter remained after filtering.

\subsection{Conventional Context Dependent Phone Models}

The initial acoustic model was trained on 650 hours of supervised
training data that comes from YouTube, Google Videos, and Broadcast
News described in~\citep{liao13asru}. The acoustic model is a 3-state
HMM with 6400 CD triphone states. This system gave us a
29.0\% word error rate on the Google Preferred test set as shown in
table~\ref{baseline125000}. By training with a sequence-level
state-MBR criterion and using a two-pass adapted decoding setup, the
best we could do was 24.0\% with a 650 hour training set. Contrast
this with adding more semi-supervised training data: at 5000 hours, we
reduced the error rate to 21.2\% for the same model size. Since we
have more data available, and models that can capture longer temporal
context, we show results for single-state CD phone
units~\citep{Senior:15}; this gives a 4\% relative improvement over
the 3-state triphone models. This type of model improves with the
amount of training data and there is little difference between CE and
CTC training criteria.

\footnotetext[1]{Asynchronous SGD gives better results with faster parameter updates.}
\begin{table}[h]
  \caption{Bidirectional-LSTM acoustic models trained on data sets of varying sizes.}
  \label{baseline125000}
  \centering
  \begin{tabular}{lllrc}
    \toprule
    Model & Training Criterion & Size & Data (hrs)   & WER(\%) \\
    \midrule
    \multirow{2}*{CD states} & CE  & 5x600 & 650    & 29.0  \\
    & CE  & 5x600 & 5000   & 21.2 \\
    \hline
    \multirow{6}*{CD phones} & CE   & 5x600 & 5000   & 20.3 \\
    & CE   & 5x600 & 50000  & 17.7 \\
    & CE   & 5x600 & 125000 & 16.7 \\
    \cline{2-5}
    & CTC  & 5x600 & 125000 & 16.5 \\ 
    & CTC, multi\_lstm\_op\footnotemark[1] & 5x600  & 125000 & 15.5 \\
    & CTC, multi\_lstm\_op\footnotemark[1] & 7x1000 & 125000 & 14.2 \\
    \bottomrule
  \end{tabular}
\end{table}

\subsection{Neural Speech Recognizer}

The entire acoustic training corpus has 1.2 billions words with a vocabulary of
1.7 million words. For the neural speech recognizer, we experimented with both spoken and written output vocabularies with the CTC loss.
For the spoken vocabulary, we decided to only model words that are seen more than 100 times.
This resulted in a vocabulary of 82473 words and an OOV rate of 0.63\%.
For the written vocabulary, we chose words seen more than 80 times, resulting in 97827 words and an OOV rate of 0.7\%.
For comparison, the full test vocabulary of our baseline has 500,000 words and an OOV rate of 0.24\%. 
We evaluated the impact of the reduced vocabulary with our best CD phone models 
and observed an increase of 0.5\% in WER (Table~\ref{ctc.words}).

\begin{figure}[t]
  \centering
  \includegraphics[width=1.0\textwidth]{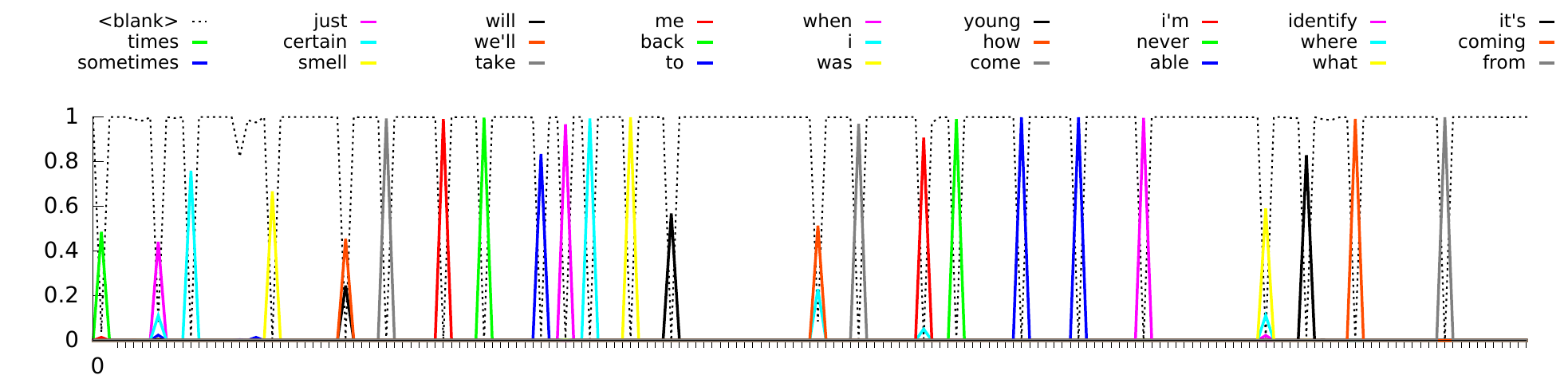}
  \caption{The word posterior probabilities as predicted by the NSR model at each time-frame (30 msec) for a segment of music video `Stressed Out' by Twenty One Pilots. We only plot the word with highest posterior and the missing words from the correct transcription:\textit{`Sometimes a certain smell will take me back to when I was young, how come I'm never able to identify where it's coming from'.}}
  \label{fig:word}
\end{figure}

In table~\ref{ctc.words} we compare CTC CD phone with CTC word
models. For both units, we trained models with $5 \times 600$ and $7
\times 1000$ bidirectional LSTM layers. As the output layer for the
word models is substantially larger, the total number of parameters
for the word models is larger than for the CD phone models for the
same number and size of LSTM layers. We also want to note that we
tried to increase the number of parameters for CD phone models, but
the reported results in table \ref{ctc.words} are the best results we
could obtain with CD phone models. As show in table \ref{ctc.words}, increasing
the number of CD phones to 35326 does not yield a reduction in error rate.
Deep decision trees tend to work mostly in scenarios when the phonetic contexts
are well matched in train and test data. Open domains such as Youtube videos
typically don't gain from a very large number of context dependent models, in
particular if temporal context is already covered with deep bidirectional LSTM
models.

As the difference in performance
between CTC and CE phone models is not too big, we ran the same
comparison for word models. As this was part of our earlier
experiments, we trained only on 50,000 hours: with CE training, the
model performed poorly with an error rate of 23.1\%, while training
with CTC loss performed substantially better at 18.7\%. This is not
unexpected as predicting longer units on a frame by frame basis with
CE makes the prediction task substantially harder. Overall, table
\ref{ctc.words} shows that the word models outperform the CD phone
models even with the handicap of a higher OOV rate for the word
models.

As mentioned earlier we can use the CTC word model directly without any
decoding or language model and the recognition output becomes the output from
the CTC layer, essentially making the CTC word model an end-to-end all-neural speech recognition model.
The entire speech recognizer becomes a single neural network.
Figure~\ref{fig:word} shows the word posterior probabilities as predicted by the model for a music video.
Even though it has not been trained on music videos, the model is quite robust and accurate in transcribing the songs.
The results are shown in the last columns in table~\ref{ctc.words} for the CTC
word models. Without any use of a language model and decoding, the CTC spoken word model has an
error rate of $14.8\%$ and the CTC written word model has $13.9\%$ WER.
The written word model is better than the conventional CD phone model which has $14.2\%$ WER obtained with decoding with a language model.
This shows that bi-directional LSTM CTC word models are capable of accurate speech
recognition with no language model or decoding involved. As a sanity check we pruned our language model heavily to a de-weighted uni-gram
model and used it with our CTC CD phone models. As expected, the error rate 
increases drastically, from $14.2\%$ to $21\%$, showing that the language model
is important for conventional models but less important for whole word CTC models.
For the spoken word model, the WER improves to $14.8\%$ when the word lattices obtained from the model are rescored with a language model.
The improvements are mostly due to conversion of spoken word forms to written forms (such as numeric entities) since the WER scoring is done in the written domain.
The WER of written word model improves only by $0.5\%$ to $13.4\%$ when the word lattices are rescored with the LM, showing the relatively small impact of the LM in the accuracy of the system.

\begin{table}[h]
  \caption{CTC CD phone models compared with CTC word models.}
  \label{ctc.words}
  \centering
  \begin{tabular}{lllllccc} \toprule
          &        &         &        &       &     & \multicolumn{2}{c}{WER(\%)} \\
    Model & Layers & Outputs & Params & Vocab & OOV(\%) & w/ LM & w/o LM \\ \midrule             
                 & 5x600  & 6400   & 14m & 500000 & 0.24 & 15.5 & ---\\ 
    CTC CD phone & 7x1000 & 6400  & 43m & 500000 & 0.24 & 14.2 & ---\\
                 & 7x1000 & 35326 & 75m & 500000 & 0.24 & 14.5 & ---\\ 
                 & 7x1000 & 6400  & 43m &  82473 & 0.63 & 14.7 & ---\\
    \hline
    \multirow{2}*{CTC spoken words}
    & 5x600  &  82473 & 57m  &  82473 & 0.63 & 14.5 & 15.8\\ 
    & 7x1000 &  82473 & 116m &  82473 & 0.63 & 13.5 & 14.8 \\
    \hline
    CTC written words & 7x1000 & 97827 & 137m & 97827 & 0.70 & 13.4 & 13.9\\     
 \bottomrule
  \end{tabular}
\end{table}

The error rate calculation disadvantages the CTC spoken word model
as the references are in written domain, but the output of the model is in spoken domain,
creating artificial errors like "three" vs "3". This is not the case for our
conventional CD phone baseline and the CTC written word model, as words are there modeled in the written domain.
To evaluate the error rate in the spoken domain, we automatically converted the
test data by force aligning the utterances with a graph built as  C * L * project(V * T),
where C is the context transducer, L the lexicon transducer, V the spoken-to-written
transducer, and T the written transcript. {\em Project} maps the input symbols to the
output symbols, thereby the output symbols of the entire graph will be in the spoken
domain. We are using the same approach to convert the written language model G
to a spoken form by calculating project(V * G) and using the spoken LM to build
the decoding graph. The word error rates in the spoken domain are shown in
table \ref{ctc.words.spoken}. The models are the same as in the previous table \ref{ctc.words}.
We can see that word models without use of any language model or decoding performs
at $12.0\%$ WER, slightly better than the CD phone model that uses an LVCSR decoder
and incorporates a 30m 5-gram language model. We can also separate the effect of
the language model from the spoken-to-written text normalization. Adding the
language model for the CTC spoken word model improves the error rate from $12.0\%$ to
$11.6\%$, showing the CTC spoken word models perform very well even without the language
model.

\begin{table}[h]
  \caption{Comparison of CD phone with spoken word models in spoken domain.}
  \label{ctc.words.spoken}
  \centering
  \begin{tabular}{lllllccc} \toprule
          &          &         &        &       &     & \multicolumn{2}{c}{Spoken WER(\%)} \\
    Model            & Layers  & Outputs& Params& Vocab  & OOV(\%) & w/ LM & w/o LM \\ \midrule
    CTC CD phone     & 7x1000  & 6400   & 43m   & 500000 & 0.24    & 12.3 & ---\\
    CTC spoken words & 7x1000  & 82473  & 116m  &  82473 & 0.63    & 11.6 & 12.0 \\ 
 \bottomrule
  \end{tabular}
\end{table}

\section{Conclusions}
\label{sec:conclusions}

We presented our Neural Speech Recognizer: an end-to-end all-neural large
vocabulary continuous speech recognizer that forgoes the use of a
pronunciation lexicon and a decoder. Mining 125,000 hours of training
data using public captions allows us to train a large and powerful
bi-directional LSTM RNN model for speech recognition with a CTC loss that predicts words.
The neural speech recognizer can model a written vocabulary of 100K words including numeric entities.
Unlike many end-to-end systems that compromise
accuracy for system simplicity, our final system performs better than
a well-trained, conventional context-dependent phone-based system
achieving a 13.4\% word error rate on a difficult YouTube video transcription task.


\small

\bibliographystyle{unsrtnat}

\begin{thebibliography}{25}
\providecommand{\natexlab}[1]{#1}
\providecommand{\url}[1]{\texttt{#1}}
\expandafter\ifx\csname urlstyle\endcsname\relax
  \providecommand{\doi}[1]{doi: #1}\else
  \providecommand{\doi}{doi: \begingroup \urlstyle{rm}\Url}\fi

\bibitem[Graves and Jaitly(2016)]{graves14icml}
Alex Graves and Navdeep Jaitly.
\newblock Towards end-to-end speech recognition with recurrent neural networks.
\newblock In \emph{Proc. ICML}, 2016.

\bibitem[Miao et~al.(2015)Miao, Gowayyed, and Metze]{miao15asru}
Yajie Miao, Mohammad Gowayyed, and Florian Metze.
\newblock {EESEN}: End-to-end speech recognition using deep {RNN} models and
  {WFST}-based decoding.
\newblock In \emph{Proc. ASRU}, 2015.

\bibitem[Bahdanau et~al.(2016)Bahdanau, Chorowski, Serdyuk, Brakel, and
  Bengio]{Bahdanu16icassp}
Dzmitry Bahdanau, Jan Chorowski, Dmitriy Serdyuk, Philemon Brakel, and Yoshua
  Bengio.
\newblock End-to-end attention-based large vocabulary speech recognition.
\newblock In \emph{Proc. ICASSP}, 2016.

\bibitem[Zhang et~al.(2016)Zhang, Pezeshki, Brakel, Zhang, Laurent, Bengio, and
  Courville]{zhang16inter}
Ying Zhang, Mohammad Pezeshki, Philémon Brakel, Saizheng Zhang, César
  Laurent, Yoshua Bengio, and Aaron Courville.
\newblock Towards end-to-end speech recognition with deep convolutional neural
  networks.
\newblock In \emph{Proc. Interspeech}, 2016.

\bibitem[{Dario Amodei, et al.}(2016)]{deepspeech2}
{Dario Amodei, et al.}
\newblock {Deep Speech 2}: End-to-end speech recognition in {English and
  Mandarin}.
\newblock In \emph{Proc. ICML}, 2016.

\bibitem[Lu et~al.(2016)Lu, Zhang, and Renals]{Liang16icassp}
Liang Lu, Xing-Xing Zhang, and Steve Renals.
\newblock On training the recurrent neural network encoder-decoder for large
  vocabulary end-to-end speech recognition.
\newblock In \emph{Proc. ICASSP}, 2016.

\bibitem[Chan et~al.(2016)Chan, Jaitly, Le, and Vinyals]{chan16ica}
William Chan, Navdeep Jaitly, Quoc~V. Le, and Oriol Vinyals.
\newblock Listen, attend and spell: A neural network for large vocabulary
  conversational speech recognition.
\newblock In \emph{Proc. ICASSP}, 2016.

\bibitem[Bahl et~al.(1991)Bahl, de~Souza, Gopalakrishnan, Nahamoo, and
  Picheny]{Bahl91}
L.R. Bahl, P.V. de~Souza, P.S. Gopalakrishnan, D.~Nahamoo, and M.A. Picheny.
\newblock Context dependent modelling of phones in continuous speech using
  decision trees.
\newblock In \emph{Proc. DARPA Speech and Natural Language Processing
  Workshop}, 1991.

\bibitem[Young et~al.(1994)Young, Odell, and Woodland]{young94}
S.J. Young, J.J. Odell, and P.C. Woodland.
\newblock Tree-based state tying for high accuracy acoustic modelling.
\newblock In \emph{Proc. ARPA Workshop on Human Language Technology}, 1994.

\bibitem[Lang et~al.(1990)Lang, Waibel, and Hinton]{lang90nn}
Kevin~J. Lang, Alex~H. Waibel, and Geoffrey~E. Hinton.
\newblock A time-delay neural network architecture for isolated word
  recognition.
\newblock \emph{Neural Networks}, 3, 1990.

\bibitem[Hochreiter and Schmidhuber(1997)]{Hochreiter:97}
Sepp Hochreiter and J\"{u}rgen Schmidhuber.
\newblock Long short-term memory.
\newblock \emph{Neural Computation}, 9\penalty0 (8):\penalty0 1735--1780,
  November 1997.
\newblock ISSN 0899-7667.
\newblock \doi{10.1162/neco.1997.9.8.1735}.

\bibitem[Graves et~al.(2006)Graves, Fern{\'a}ndez, Gomez, and
  Schmidhuber]{Graves:06}
Alex Graves, Santiago Fern{\'a}ndez, Faustino Gomez, and J{\"u}rgen
  Schmidhuber.
\newblock Connectionist temporal classification: {Labelling} unsegmented
  sequence data with recurrent neural networks.
\newblock In \emph{Proc. ICML}, 2006.

\bibitem[Sak et~al.(2015{\natexlab{a}})Sak, Senior, Rao, and Beaufays]{Sak:15b}
Ha{\c{s}}im Sak, Andrew Senior, Kanishka Rao, and Fran{\c{c}}oise Beaufays.
\newblock Fast and accurate recurrent neural network acoustic models for speech
  recognition.
\newblock \emph{arXiv preprint arXiv:1507.06947}, 2015{\natexlab{a}}.

\bibitem[Schuster and Paliwal(1997)]{Schuster:97}
Mike Schuster and Kuldip~K. Paliwal.
\newblock Bidirectional recurrent neural networks.
\newblock \emph{Signal Processing, IEEE Transactions on}, 45\penalty0
  (11):\penalty0 2673--2681, 1997.
\newblock ISSN 1053-587X.
\newblock \doi{10.1109/78.650093}.

\bibitem[Sak et~al.(2015{\natexlab{b}})Sak, Senior, Rao, Irsoy, Graves,
  Beaufays, and Schalkwyk]{Sak:15a}
Ha{\c{s}}im Sak, Andrew Senior, Kanishka Rao, Ozan Irsoy, Alex Graves,
  Fran{\c{c}}oise Beaufays, and Johan Schalkwyk.
\newblock Learning acoustic frame labeling for speech recognition with
  recurrent neural networks.
\newblock In \emph{Proc. ICASSP}, 2015{\natexlab{b}}.

\bibitem[Sak et~al.(2013)Sak, Beaufays, Nakajima, and Allauzen]{Sak:13}
Ha{\c{s}}im Sak, Fran{\c{c}}oise Beaufays, Kaisuke Nakajima, and Cyril
  Allauzen.
\newblock Language model verbalization for automatic speech recognition.
\newblock In \emph{Proc. ICASSP}, 2013.

\bibitem[Dean et~al.(2012)Dean, Corrado, Monga, Chen, Devin, Le, Mao, Ranzato,
  Senior, Tucker, Yang, and Ng]{Dean:12}
Jeffrey Dean, Greg Corrado, Rajat Monga, Kai Chen, Matthieu Devin, Quoc~V. Le,
  Mark~Z. Mao, Marc'Aurelio Ranzato, Andrew~W. Senior, Paul~A. Tucker, Ke~Yang,
  and Andrew~Y. Ng.
\newblock Large scale distributed deep networks.
\newblock In \emph{NIPS}, 2012.

\bibitem[Sak et~al.(2014)Sak, Senior, and Beaufays]{Sak:14a}
Hasim Sak, Andrew Senior, and Francoise Beaufays.
\newblock {Long Short-Term Memory Recurrent Neural Network Architectures for
  Large Scale Acoustic Modeling}.
\newblock In \emph{Proc. Interspeech}, 2014.

\bibitem[aut()]{autocaps}
Automatic captions in {YouTube}.
\newblock
  \url{http://googleblog.blogspot.com/2009/11/automatic-captions-in-youtube.html}.
\newblock Online: accessed Oct 12, 2016.

\bibitem[ytp()]{ytpref}
Google preferred lineup explorer - {YouTube}.
\newblock \url{http://youtube.com/yt/lineups/}.
\newblock Online: accessed March, 2016.

\bibitem[Kuznetsov et~al.(2016)Kuznetsov, Liao, Mohri, Riley, and
  Roark]{vitaly16inter}
Vitaly Kuznetsov, Hank Liao, Mehryar Mohri, Michael Riley, and Brian Roark.
\newblock Learning {N-gram} language models from uncertain data.
\newblock In \emph{Proc. Interspeech}, 2016.

\bibitem[Panayotov et~al.(2015)Panayotov, Chen, Povey, and
  Khudanpur]{vassil15ica}
Vassil Panayotov, Guoguo Chen, Daniel Povey, and Sanjeev Khudanpur.
\newblock {LibriSpeech}: an {ASR} corpus based on public domain audio books.
\newblock In \emph{Proc. ICASSP}, 2015.

\bibitem[Kapralova et~al.(2014)Kapralova, Alex, Weinstein, Moreno, and
  Siohan]{olga14inter}
Olga Kapralova, John Alex, Eugene Weinstein, Pedro Moreno, and Olivier Siohan.
\newblock A big data approach to acoustic model training corpus selection.
\newblock In \emph{Proc. Interspeech}, 2014.

\bibitem[Liao et~al.(2013)Liao, McDermott, and Senior]{liao13asru}
Hank Liao, Erik McDermott, and Andrew Senior.
\newblock Large scale deep neural network acoustic modeling with
  semi-supervised training data for youtube video transcription.
\newblock In \emph{Proc. ASRU}, 2013.

\bibitem[Senior et~al.(2015)Senior, Sak, and Shafran]{Senior:15}
A.~Senior, H.~Sak, and I.~Shafran.
\newblock Context dependent phone models for {LSTM} {RNN} acoustic modelling.
\newblock In \emph{Proc. ICASSP}, 2015.

\end{thebibliography}

\end{document}